\documentclass{article}
\usepackage{microtype}

\usepackage{tcolorbox}
\usepackage{xcolor}
\usepackage{amsmath,amssymb}
\usepackage{algorithmic}
\usepackage{algorithm}
\usepackage{bm}
\usepackage{booktabs}
\usepackage{comment}
\usepackage{multirow}
\usepackage{framed}

\usepackage{url}
\makeatletter
\newcommand{\figcaption}[1]{\def\@captype{figure}\caption{#1}}
\newcommand{\tblcaption}[1]{\def\@captype{table}\caption{#1}}
\makeatother

\newcount\K
\def\bla#1{
\K=0 \loop\ifnum\K<#1
{\textcolor[gray]{0.9}{{\it bla bla bla bla bla bla bla bla bla bla bla bla bla bla bla}}}
\advance\K by1\repeat
}

\newcommand{\todo}[1]{
\ifx#10
\textcolor{red}{$0.00_{\pm 0.00}$}
\else
\textcolor{red}{#1}
\fi
}

\usepackage{pifont}
\usepackage{spconf,amsmath,graphicx}
\usepackage{amssymb}
\usepackage{booktabs}
\usepackage[capitalize]{cleveref}
\crefname{section}{Sec.}{Secs.}
\Crefname{section}{Section}{Sections}
\Crefname{table}{Table}{Tables}
\crefname{table}{Tab.}{Tabs.}
\title{Pyramid Coder: Hierarchical Code Generator for \\ Compositional Visual Question Answering}
\name{Ruoyue Shen, Nakamasa Inoue, Koichi Shinoda}
\address{Tokyo Institute of Technology}
\begin{document}
\maketitle
\begin{abstract}
Visual question answering (VQA) is the task of providing accurate answers to natural language questions based on visual input. Programmatic VQA (PVQA) models have been gaining attention recently. These use large language models (LLMs) to formulate executable programs that address questions requiring complex visual reasoning.
However, there are challenges in enabling LLMs to comprehend the usage of image processing modules and generate relevant code. To overcome these challenges, this paper introduces PyramidCoder, a novel prompting framework for PVQA models. PyramidCoder consists of three hierarchical levels, each serving a distinct purpose: query rephrasing, code generation, and answer aggregation. 
Notably, PyramidCoder utilizes a single frozen LLM and pre-defined prompts at each level, eliminating the need for additional training and ensuring flexibility across various LLM architectures. 
Compared to the state-of-the-art PVQA model, our approach improves accuracy by at least 0.5\% on the GQA dataset, 1.4\% on the VQAv2 dataset, and 2.9\% on the NLVR2 dataset.

\end{abstract}

\section{Introduction}
\label{sec:intro}
Visual question answering (VQA), which aims to provide accurate answers to natural language questions based on visual input, is a crucial research topic in computer vision and natural language processing~\cite{Hudson2019GQADataset, goyal2017vqav2, suhr2019nlvr2}. The last decade has seen significant advances in deep learning applied to VQA, with the development of end-to-end multimodal models such as GLIP~\cite{Li2022GLIP} and PNP-VQA~\cite{tiong2022pnpvqa}. However, despite these advances, compositional VQA, which involves complex spatial relationships and object attributes, remains difficult due to the lack of an explicit understanding of visual elements during the inference process.

To address the difficulties in compositional VQA, recent studies~\cite{Suris2023ViperGPT, Gupta2022VisProg, Subramanian2023CodeVQA} have proposed models that generate executable programs to answer questions based on given images, questions, and image processing modules. These models, which we refer to as Programmatic VQA (PVQA) models, represent a novel approach to integrating multimodal inputs, leading to more tractable and traceable inference.
Typically, PVQA models consist of three components: a large language model (LLM) for code generation, a set of image processing modules, and a Python executor. The LLM analyzes a given query (text-form question) and generates Python code to answer it using a predefined API set of image processing modules, including both low-level modules such as image cropping and high-level modules such as object detection. The generated code is then executed by the Python executor to produce the answer. This architecture allows questions to be decomposed into manageable segments of code, providing a more precise approach to compositional VQA.
However, enabling the LLM to understand API usage and generate appropriate code for VQA is not always straightforward.
\begin{figure}
\centering \includegraphics[width=1.0\linewidth]{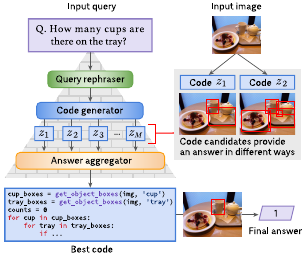}
\caption{
PyramidCoder is a hierarchical framework comprising three modules. It generates multiple code candidates, maximizing the diversity of solutions available for a given input query to produce the correct answer effectively.
}
\label{fig:toppage}
\end{figure}

To fully leverage the capabilities of LLMs, a number of prompting frameworks have been developed for natural language processing tasks. Chain-of-Thought (CoT)~\cite{Wei2022ChainOfThought_CoT}, for example, demonstrates how structured prompts can guide LLMs to decompose complex problems into a series of simpler, linked steps. There are also extensions of CoT such as Tree of Thoughts~\cite{Yao2023TreeOfThoughts_ToT}, which creates a tree structure of thinking steps. These frameworks not only improve performance, but also provide greater transparency into their reasoning process.

Inspired by these approaches, this paper introduces PyramidCoder, a novel code generation prompting framework for PVQA models. 
Previous PVQA models~\cite{ Suris2023ViperGPT, Gupta2022VisProg, Subramanian2023CodeVQA} employ Input-Output (IO) prompting to generate code from prompt input directly. However, IO prompting constraints a singular solution for a question, resulting in a simplistic and uniform code generation. In contrast, PyramidCoder explores the diversity of question statements and possible solutions through a hierarchical framework consisting of three levels. 
As shown in Figure~\ref{fig:toppage}, the first level rephrases a given query into multiple variations. The second level then generates multiple code candidates corresponding to the rephrased queries. Finally, the third level aggregates them and generates the final code and answer for the input query.
All these procedures are implemented with a single frozen LLM and pre-defined prompts at each level, without the need for additional training. In experiments, we substantiate the efficacy of PyramidCoder across GQA~\cite{Hudson2019GQADataset}, VQAv2~\cite{goyal2017vqav2}, and NLVR2~\cite{suhr2019nlvr2} datasets, showing its ability to significantly enhance VQA performance.
In summary, our contribution is three-fold:
{
\setlength{\leftmargini}{16pt}
\begin{enumerate}
\setlength{\itemsep}{4pt}
\setlength{\parskip}{3pt}
\item We introduce PyramidCoder, a novel code generation framework for PVQA consisting of three modules: a query rephraser, a code generator, and an answer aggregator.
\item We propose a prompting method to implement all modules with a single frozen LLM to avoid additional training. This approach offers flexibility and is not constrained by the specific LLM employed.
\item We demonstrate the effectiveness of our method over the state-of-the-art CodeVQA~\cite{Subramanian2023CodeVQA} model by conducting experiments on three VQA datasets.
\end{enumerate}
}
\section{Related work}
\label{sec:related_work}

\subsection{Visual Question Answering}
VQA is an interdisciplinary research area combining computer vision and natural language processing. It aims to develop AI systems that respond to queries about images by integrating multimodal information. Early VQA models relied on convolutional neural networks to extract image features and recurrent neural networks to process textual inputs~\cite{malinowski2015ask, huang2021diagnostic, zhang2022context}. The introduction of attention mechanisms 
~\cite{yang2016stackedattention, anderson2018bottomupandtopdown, sarkar2022grad, wu2022ques, Shen_2023_WACV} has since improved these models’ capacity to handle complex questions and comprehend fine-grained image details. Recently, multimodal pretaining approaches  
~\cite{Li2022GLIP, li2021unsupervised, le2021vttransformer} have been employed, leveraging large-scale pretraining on both image and text data to capture richer contextual information.

However, fine-tuning multimodal pre-trained models for VQA requires substantial expertise, vast amounts of data, and significant computational resources.
PVQA models~\cite{Suris2023ViperGPT, Gupta2022VisProg, Subramanian2023CodeVQA} address these issues by utilizing an LLM to generate a Python-like code, which is then executed with predefined APIs to find answers to questions.
These models can quickly adapt to new tasks or domains with minimal data, without the need for additional training or fine-tuning. 
Despite these advantages, PVQA models using IO prompting display a level of simplicity that hampers the effective utilization of LLM capabilities. Consequently, methods for activating the latent potential of LLMs and guiding them towards optimal API utilization remain an underexplored domain.

\subsection{Large Language Models}
\noindent{\textbf{Code Generation Models.}}
Large language models have demonstrated remarkable capabilities in a wide range of natural language processing tasks. While general LLMs find application in various tasks~\cite{brown2020gpt3, touvron2023llama}, a subset of these models has been specifically tailored for code generation. This specialized category of LLMs undergoes training on extensive corpora of programming-related text, enabling them to comprehend and generate code in response to natural language descriptions. Codex~\cite{Chen2021Codex}, an extension of the GPT-3 series, demonstrates proficiency in over a dozen languages.
StarCoder~\cite{Li2023StarCoder}, as one of the pioneering open-source Code LLMs, undergoes training on datasets sourced from Github licensed data, covering over 80 programming languages. CodeLlama~\cite{roziere2023codellama}, fine-tuned from Llama2 with a higher sampling of code, augments support for larger input contexts and zero-shot instruction following ability. Noteworthy among recent developments are native multimodality LLMs such as Gemini~\cite{team2023gemini}. These models leverage multimodal data (i.e., text, images, and code) throughout the entirety of the training process to enhance their proficiency in generating code that meets specific requirements.

\noindent{\textbf{Prompting.}}
Prompting methods are employed to optimize prompts for LLMs in order to maximize their capacity. Chain-of-Thought (CoT)~\cite{Wei2022ChainOfThought_CoT} guides LLMs in generating intermediate reasoning steps before predicting the answer. This method has inspired several extensions.
Notably, Self-consistency~\cite{Wang2023SelfConsistency} samples from the LLM decoder multiple times and aggregates the final answer through majority voting. AutoCoT~\cite{Zhang2023AutomaticCoT} automatically constructs demonstrations through question clustering and zero-shot CoT reasoning chain generation. ReAct~\cite{Yao2022ReAct} alternately generates thoughts and actions, performs the action based on the thought, and adjusts the thought based on the action result. Similar to CoT prompting and its variations, Tree of Thoughts (ToT)~\cite{Yao2023TreeOfThoughts_ToT} decomposes problems into smaller thoughts and explores multiple solution paths in parallel, forming a tree structure of thoughts.

\newcommand{\qr}[0]{R}
\newcommand{\cg}[0]{G}
\newcommand{\ag}[0]{A}
\begin{figure*}
\centering
\includegraphics[width=1.0\linewidth]{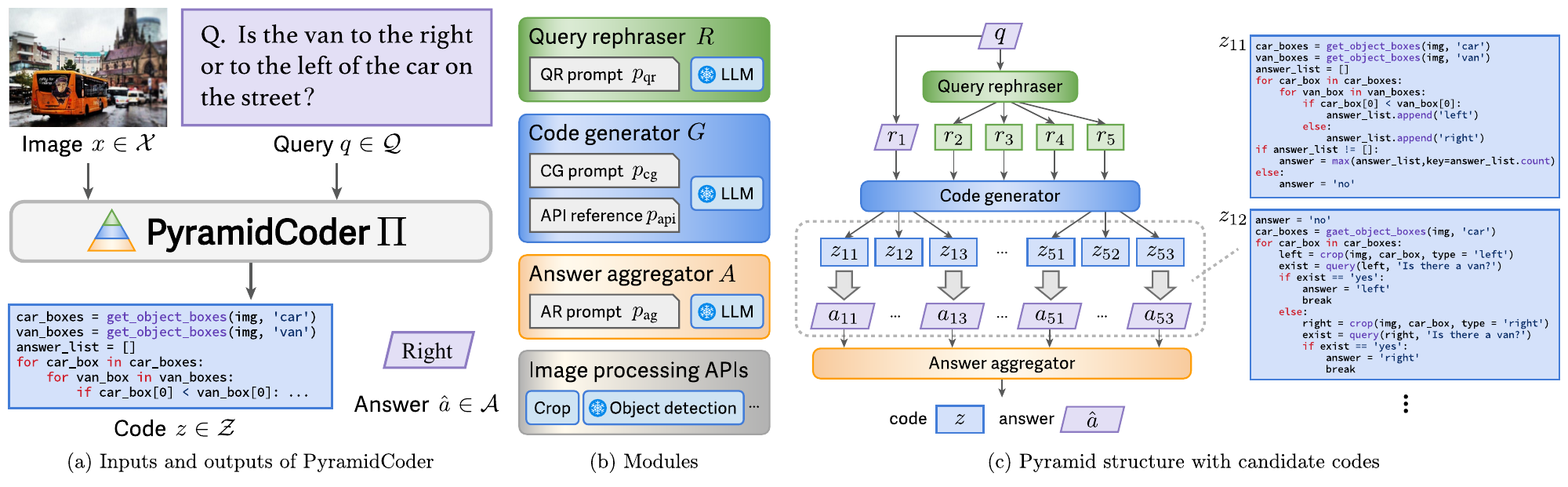}
\caption{\textbf{Framework overview.} 
    (a) PyramidCoder $\Pi$ takes an image $x$ and a query $q$ as input, producing both the code and answer for the input image-query pair. (b) PyramidCoder contains three modules: a query rephraser $\qr$, a code generator $\cg$, and an answer aggregator $\ag$. Image processing APIs are also used in the code-executing process. (c) For each input query, the PyramidCoder generates diverse expressions $r_{i}$ of the input query and multiple code candidates $z_{ij}$, each representing a distinct solution to the given query. Finally, these codes $z_{ij}$ and their corresponding pre-execution results $a_{ij}$ are aggregated to derive the final answer $\hat{a}$ and the accompanying interpretable code $z$.
    }
\label{fig:overview}
\end{figure*}

\section{Proposed Method}
This section introduces PyramidCoder, a hierarchical code generation framework for PVQA. As shown in Figure~\ref{fig:overview}, PyramidCoder generates a Python code to answer a question given a query, an image, and a set of pre-defined APIs for simple processing tasks. The framework consists of three modules: a query rephraser $\qr$, a code generator $\cg$, and an answer aggregator $\ag$. These three modules are implemented with a single frozen LLM and work in conjunction to enhance the quality of code generation within the pyramid structure.

\subsection{Overall Framework}
\label{framework}

\noindent \textbf{Problem setting.}
We follow the notation used in previous studies~\cite{ Suris2023ViperGPT, Subramanian2023CodeVQA, Johnson2017Inferring}.
Let $x \in \mathcal{X}$ be an input image and $q \in \mathcal{Q}$ be a textual query, where $\mathcal{X}$ is a set of images and $\mathcal{Q}$ is a set of textual queries.
The goal is to generate a code $z \in \mathcal{Z}$ that returns the answer $a \in \mathcal{A}$ to the query given the image, where $\mathcal{Z}$ is a set of executable codes and $\mathcal{A}$ is a set of answers. This study assumes that $\mathcal{Z}$ is a set of Python codes. Note that the three sets $\mathcal{Q}, \mathcal{Z}$ and $\mathcal{A}$ are subsets of the whole set of texts, which we denote as $\mathcal{T}$. We use a frozen LLM $\pi : \mathcal{T} \to \mathcal{T}$ to find mappings between these subsets.

\noindent \textbf{Framework.}
The PyramidCoder framework consists of two stages: the coding stage and the execution stage.
The coding stage generates a code as
\begin{align}
z = \Pi (q, x),
\end{align}
where
$\Pi : \mathcal{Q} \times \mathcal{X} \to \mathcal{Z}$
is the code generation process. 
In the execution stage, the generated code is executed with the input image to obtain an answer as
\begin{align}
a = \Lambda(x, z),
\end{align}
where $a$ is the predicted answer and $\Lambda : \mathcal{X} \times \mathcal{Z} \to \mathcal{A}$ is the Python execution engine.
This paper studies the design of the code generation process $\Pi$.

\noindent \textbf{Algorithm.}
The proposed code generation process $\Pi$ is summarized in Algorithm~\ref{alg1}.
It consists of three steps.
First, given an input query $q \in \mathcal{Q}$,
the query rephraser $R$ is applied to obtain a set of rephrased queries $\{r_{i}\}_{i=1}^{N}$.
Here, each $r_{i}$ is a rephrased version of $q$, intended to make coding more effective with a more comprehensive understanding of the input query.
Second, the code generator $G$ is applied to each rephrased query. This step produces multiple candidate codes $\{z_{ij}\}_{j=1}^{M}$ for each rephrased query $r_{i}$ and pre-executes them to obtain candidate answers $a_{ij} = \Lambda(x, z_{ij})$.
Finally, the answer aggregator $A$ is applied to the set of candidate code-answer pairs. This step chooses the final answer $a_{\tau}$ and the best code $z_{\tau}$, where $\tau$ is the index of the chosen pair.
As shown in Figure~\ref{fig:overview}, 
PyramidCoder implements all modules with a single frozen LLM, and the code generation process is represented by a pyramid structure of texts (queries and codes) generated by the LLM.

\subsection{Module implementation with a single LLM}
\label{modules}

Given a frozen LLM $\pi : \mathcal{T} \to \mathcal{T}$ pre-trained on a large text dataset comprising both natural language and programming code, we implement the query rephraser $R$, the code generator $G$ and the answer aggregator $A$ with the aforementioned LLM without any additional training.
Below, we describe the motivation and definition of each module.
\begin{algorithm}[t]
\caption{PyramidCoder $\Pi(q, x)$}
\label{alg1}
\begin{algorithmic}
\REQUIRE Query $q \in \mathcal{Q}$, Image $x \in \mathcal{X}$, LLM $\pi$
\ENSURE Code $z \in \mathcal{Z}$, Answer $\hat{a} \in \mathcal{A}$
\STATE $Z \leftarrow \emptyset$
\STATE $[r_{1}, r_{2}, \cdots, r_{N}] \leftarrow R(q)$ \# Query rephraser
\FOR{$i = 1, \cdots, N$}
\STATE $[z_{i1}, z_{i2}, \cdots, z_{iM}] \leftarrow G(r_{i})$ \# Code generator
\FOR{$j = 1, \cdots, M$}
\STATE $a_{ij} \leftarrow \Lambda(x, z_{ij})$ \# Python executor
\STATE $Z \leftarrow Z \cup \{(z_{ij}, a_{ij})\}$
\ENDFOR
\ENDFOR
\STATE $\bm{\sigma} \leftarrow A_{\text{ans}}(Z)$  \# Answer aggregator
\STATE $\tau \leftarrow A_{\text{code}}(Z_{\bm{\sigma}})$  \# Answer aggregator
\RETURN Code $z = z_{\tau}$, Answer $\hat{a} = a_{\tau}$
\end{algorithmic}
\end{algorithm}

\subsubsection{Query Rephraser}
\label{qr}
\noindent \textbf{Motivation.} 
The query rephraser module in PyramidCoder is designed to enhance the interpretability and robustness of the model's comprehension of input queries.
The VQA task demands a precise understanding of diverse query expressions, as users may articulate questions in various linguistic forms. It is widely recognized that the input prompt, including input queries, significantly influences the response quality of LLMs. The query rephraser module dynamically reformulates an input query into diverse expressions while preserving the underlying semantics. Consequently, the subsequent code generation stage benefits from a more comprehensive and adaptable understanding of user queries, thereby alleviating potential misinterpretations and improving overall task performance.

\begin{figure}
\centering
\includegraphics[width=\linewidth]{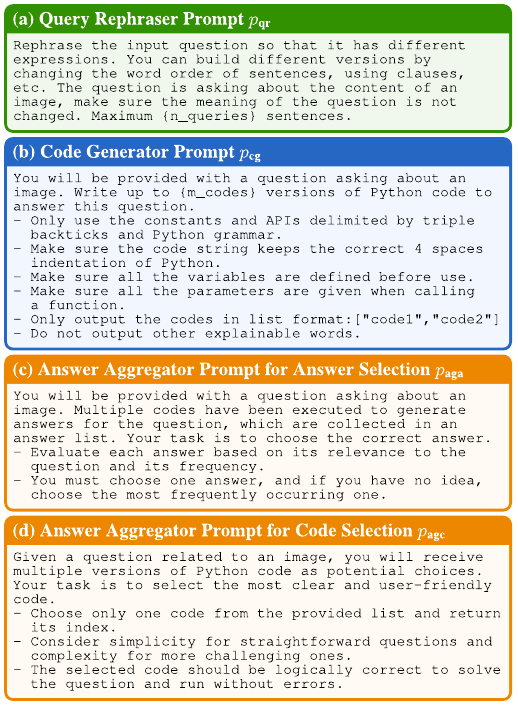}
\caption{Prompt definitions.}
\label{fig:prompt}
\end{figure}

\noindent \textbf{Definition.} We define the query rephraser by
\begin{align}
R(q) \triangleq \pi( p_{\text{qr}} + s_{\text{qr}} + q ),
\end{align}
where $q$ is an input query, $p_{\text{qr}}$ is the query rephraser prompt as shown in Figure~\ref{fig:prompt}a, and $s_{\text{qr}}$ is the few-shot examples prompt used for rephrasing.
The operation $+$ indicates the concatenation of texts.

\begin{table*}[t]
\centering
\setlength{\tabcolsep}{9.00pt}
\caption{Comparison with SOTA methods for compositional visual question answering.
Accuracy (\%) on three datasets is reported.
Code: $\checkmark$ indicates code generation methods. LLM: Model names of large language models. 
}
\vspace{7pt}
\begin{tabular}{l|c|c|cccc}
\toprule
\multirow{2}{*}{Method} & \multirow{2}{*}{Code} & \multirow{2}{*}{LLM} & GQA & GQA & VQAv2 & NLVR2 \\
& & & val2000 & testdev & val4000  & test \\
\midrule
CodeVQA~\cite{Subramanian2023CodeVQA} & $\checkmark$ & StarCoder-Base & 46.5 & 41.0 & 56.4 & 53.5 \\
PyramidCoder (Ours)  & $\checkmark$ & StarCoder-Base & \textbf{48.2} & \textbf{41.5} & \textbf{57.8} &  \textbf{59.2} \\
\midrule
CodeVQA~\cite{Subramanian2023CodeVQA} & $\checkmark$ & CodeLlama-7b-Python & 48.0 & 42.8 & 61.3 & 57.8 \\
PyramidCoder (Ours) & $\checkmark$ & CodeLlama-7b-Python & \textbf{51.3} & \textbf{43.4} & \textbf{62.8} & \textbf{60.7} \\
\midrule
\textcolor{gray}{Few-shot PnP-VQA~\cite{tiong2022pnpvqa}} & \textcolor{gray}{$-$} & \textcolor{gray}{$-$} & \textcolor{gray}{55.1} & \textcolor{gray}{46.6} & \textcolor{gray}{66.8} & \textcolor{gray}{63.4} \\
\bottomrule
\end{tabular}
\label{tab:sota}
\end{table*}

\subsubsection{Code Generator}
\label{cg}
\noindent \textbf{Motivation.}
The development of the code generator module is driven by the recognition that complex problem-solving often involves an inherent diversity and multiplicity of solutions. Traditional PVQA models~\cite{ Suris2023ViperGPT, Gupta2022VisProg, Subramanian2023CodeVQA} predominantly focus on generating a singular Python code for a given input query. However, this conventional methodology overlooks the intrinsic variability in problem-solving strategies and fails to capture the diversity of viable solutions that may lead to the correct answer. By introducing the capability to generate multiple Python codes for a single input query, the code generator aims to make full use of potential solutions associated with a given query. This approach is grounded in the understanding that various coding strategies may be employed to achieve the same desired outcome, fostering a more nuanced and adaptable code generation in complex problem-solving domains.

\begin{figure}
\centering
\includegraphics[width=\linewidth]{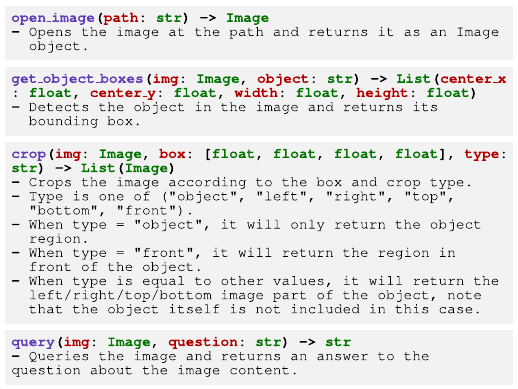}
\caption{API reference $p_{\text{api}}$.}
\label{fig:api}
\end{figure}

\noindent \textbf{Definition.}
We define the code generator by
\begin{align}
G(r) \triangleq \pi( p_{\text{cg}}  + p_{\text{api}} + s_{\text{cg}} + r ),
\end{align}
where $r$ is a rephrased query, $p_{\text{cg}}$ is the code generator prompt in Figure~\ref{fig:prompt}b, $p_{\text{api}}$ is API reference texts detailed in Figure~\ref{fig:api}, and $s_{\text{cg}}$ is the few-shot examples prompt for code generation.

\subsubsection{Answer Aggregator}
\label{cg}
\noindent \textbf{Motivation.}
The traditional approach often employs majority voting to aggregate solutions. 
However, this approach encounters challenges when failed code execution leads to the default answer being the majority or when multiple answers achieve majority status. In response to these limitations, the answer aggregator utilizes the capabilities of LLMs to select the final output, considering not only the frequency but also the semantic compatibility between the query and potential answers. 
The answer aggregator operates in two steps. First, selecting the final answer from the candidate answer set, and then selecting the most suitable code that produced this final answer during pre-execution.
This sequential approach minimizes input token consumption and alleviates potential mismatches between answers and their corresponding code that might be caused by hallucinations.

\noindent \textbf{Definition.}
Given a set of candidate answers and codes $Z = \{(z_{k}, a_{k})\}_{k=1}^{O}$, we define the answer aggregator by
\begin{align}
A_{\text{ans}}(Z) &\triangleq \pi( p_{\text{aga}} + [a_{1}, a_{2}, \cdots, a_{O} ] ), \\
A_{\text{code}}(Z_{\bm{\sigma}}) &\triangleq \pi( p_{\text{agc}} + [z_{\sigma_{1}}, z_{\sigma_{2}}, \cdots, z_{\sigma_{l}} ] ),
\end{align}
where $p_{\text{aga}}$ is the prompt used by the answer aggregator to select the final answer shown in Figure~\ref{fig:prompt}c, and $p_{\text{agc}}$ is the prompt to choose the best code shown in Figure~\ref{fig:prompt}d. The operation [] represents the aggregation of elements into a list and the conversion to string format. $Z_{\bm{\sigma}} = \{z_{\sigma_{l}}\}_{l=1}^{L}$ is a subset of $Z$, where $\bm{\sigma} = \{\sigma_{l}\}_{l=1}^{L}$ is the output of $A_{\text{ans}}$ and represents the set of code indices where the pre-execution output for the code matches the selected answer.

\section{Experiments}

\subsection{Experimental Settings}
\noindent {\bf Datasets and evaluation metrics.} 
Three VQA datasets, namely GQA~\cite{Hudson2019GQADataset}, VQAv2~\cite{goyal2017vqav2}, and NLVR2~\cite{suhr2019nlvr2}, are employed for assessment in the conducted experiment.
The GQA dataset is known for its challenging questions that involve spatial reasoning, commonsense reasoning, and an understanding of image details. 
The VQAv2 dataset comprises images from the COCO dataset paired with open-ended questions. The questions exhibit a range of complexities and topics, and answers are expected to be provided in free-form text. The NLVR2 dataset consists of statements about pairs of images, each annotated as true or false based on the image content. For our experiments, these statements are reformulated into a query format.

Following the baseline method, 12 in-context examples are sampled to construct the in-context prompt for the GQA and VQAv2 datasets, while 6 examples are used for the NLVR2 dataset. To conserve computational resources, for certain datasets, we utilize the same randomly sampled subset as the baseline for evaluation purposes. In the results table, GQA val2000 consists of a subset of 2000 samples randomly selected from the GQA validation set, and VQAv2 val4000 comprises 4000 examples randomly sampled from the VQAv2 validation set. The evaluation metric is the exact match accuracy for case-insensitive answers.

\subsection{Implementation Details}
\noindent {\bf Frozen LLM $\pi$.}
PyramidCoder exhibits insensitivity to the underlying base LLM model $\pi$. In our evaluation, we employ one closed general LLM, namely, GPT-3.5 (\texttt{\small gpt-3.5-turbo}) available through the OpenAI API, and two open-source Code LLMs: StarCoderBase~\cite{Li2023StarCoder} and CodeLlama-7b-Python~\cite{roziere2023codellama}.

\noindent {\bf Framework and APIs.}
We validate the efficacy of our approach by employing the state-of-the-art PVQA model, CodeVQA~\cite{Subramanian2023CodeVQA}, as our baseline framework. Utilizing identical image processing modules as those employed in CodeVQA, we use GroundingDINO\cite{liu2023groundingdino} for object detection through \texttt{get\_object\_boxes()}. For image captioning, we utilize BLIP~\cite{li2022blip} within \texttt{query()}. Additionally, the question-answering module incorporated within \texttt{query()} employs the same LLM as the one used in the code generation stage.

\begin{table}
\centering
\caption{Comparison with state-of-the-art prompting methods.}
\vspace{7pt}
\setlength{\tabcolsep}{9.00pt}
\begin{tabular}{l|c|c}
\toprule
Method & GQA & VQAv2 \\
\midrule
Input-Output (Default) & 48.0 & 61.3\\
Chain-of-Thought~\cite{Wei2022ChainOfThought_CoT}& 48.3 & 61.0\\
Tree of Thoughts~\cite{Yao2023TreeOfThoughts_ToT} & 48.7 & 61.5\\
PyramidCoder (Ours) & \textbf{51.3} & \textbf{62.8}\\
\bottomrule
\end{tabular}
\label{tab:prompting}
\end{table}

\begin{table}
\centering
\caption{Results using different LLMs on GQA val2000. IO is the default strategy used in CodeVQA.}
\vspace{7pt}
\setlength{\tabcolsep}{9pt}
\begin{tabular}{l|cc}
\toprule
LLM & IO & Ours \\
\midrule
gpt3.5-turbo & 46.6 & \textbf{55.8} \\
StarCoder-Base & 46.5 & \textbf{48.2} \\
CodeLlama-7b & 48.0 & \textbf{51.3}  \\
\bottomrule
\end{tabular}
\label{tab:llm}
\end{table}

\begin{table}
\centering
\caption{Ablation study with respect to the three modules.}
\vspace{7pt}
\setlength{\tabcolsep}{9.00pt}
\begin{tabular}{l|c|c}
\toprule
Method & GQA & VQAv2 \\
\midrule
PyramidCoder & 51.3 & 62.8\\
w/o query rephraser & 47.8 & 61.0 \\
w/o code generator & 49.4 & 61.0\\
w/o answer aggregator & 51.0 & 62.1\\
\bottomrule
\end{tabular}
\label{tab:ablation}
\end{table}

\subsection{Results and Analysis}
\noindent{\textbf{VQA performance.}} 
Table~\ref{tab:sota} lists the performance across three VQA datasets. Our baseline is the state-of-the-art CodeVQA model, which uses default IO prompting. The same settings and APIs are utilized for all experiments. Compared with the baseline, PyramidCoder demonstrates relative improvements of a minimum of 1.65 points on the GQA val2000 set, 0.5 points on the GQA test set, 1.4 points on the VQAv2 val4000 set, and 2.9 points on the NLVR2 test set. These results indicate the effectiveness of  PyramidCoder, which maintains its performance across a variety of underlying LLMs and datasets.

\noindent{\textbf{Prompting Methods.}} 
Table~\ref{tab:prompting} provides a comparative analysis of our PyramidCoder against existing prompting methodologies, namely few-shot CoT and few-shot ToT with a creative writing setup. All methods employ identical in-context examples and few-shot configurations as described earlier. While the preceding prompting techniques show improvements in comparison to default IO prompting, they tend to generate homogeneous codes. In contrast, PyramidCoder surpasses these methods by producing diverse code candidates, which provides more substantial reasoning evidence.

\begin{figure}
\centering \includegraphics[width=1.0\linewidth]{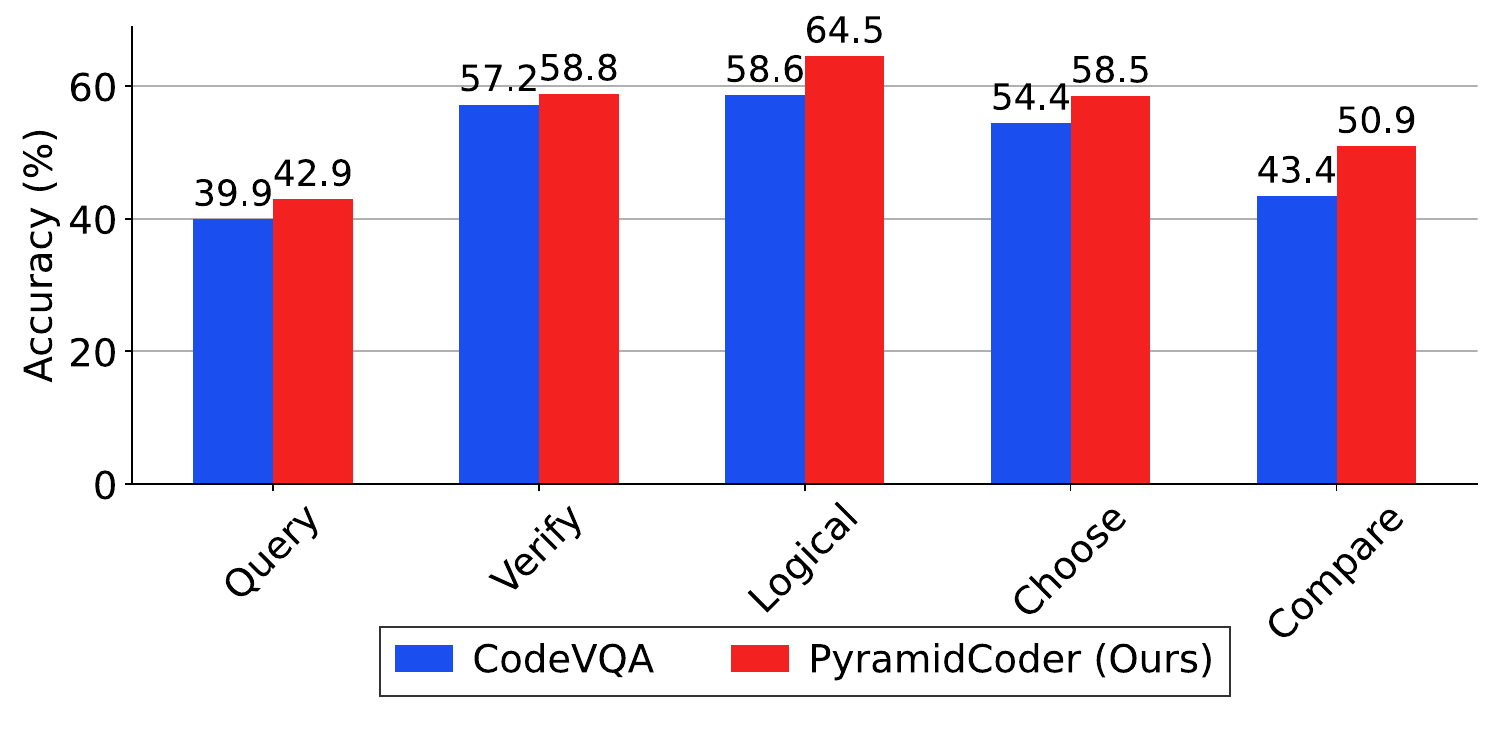}
\caption{QA Accuracy by question type.}
\label{fig:questiontype}
\end{figure}

\noindent{\textbf{Analysis by question type.}}
Figure~\ref{fig:questiontype} illustrates the QA accuracy across diverse question types within the GQA val2000 dataset. 
The improvements are particularly evident in addressing logical questions involving logical inference, choose questions characterized by the inclusion of alternative options within the query, and compare questions requiring comparisons between two or more objects. These findings reveal the model's proficiency in handling queries necessitating intricate reasoning and comprehension of visual content. Moreover, PyramidCoder demonstrates heightened performance even in addressing simpler question types, such as verify questions that require binary responses (yes/no) and queries involving open-ended questions. This comprehensive enhancement underscores the versatility and efficacy of the PyramidCoder in diverse question-answering scenarios.

\begin{figure*}
\centering
\includegraphics[width=\linewidth]{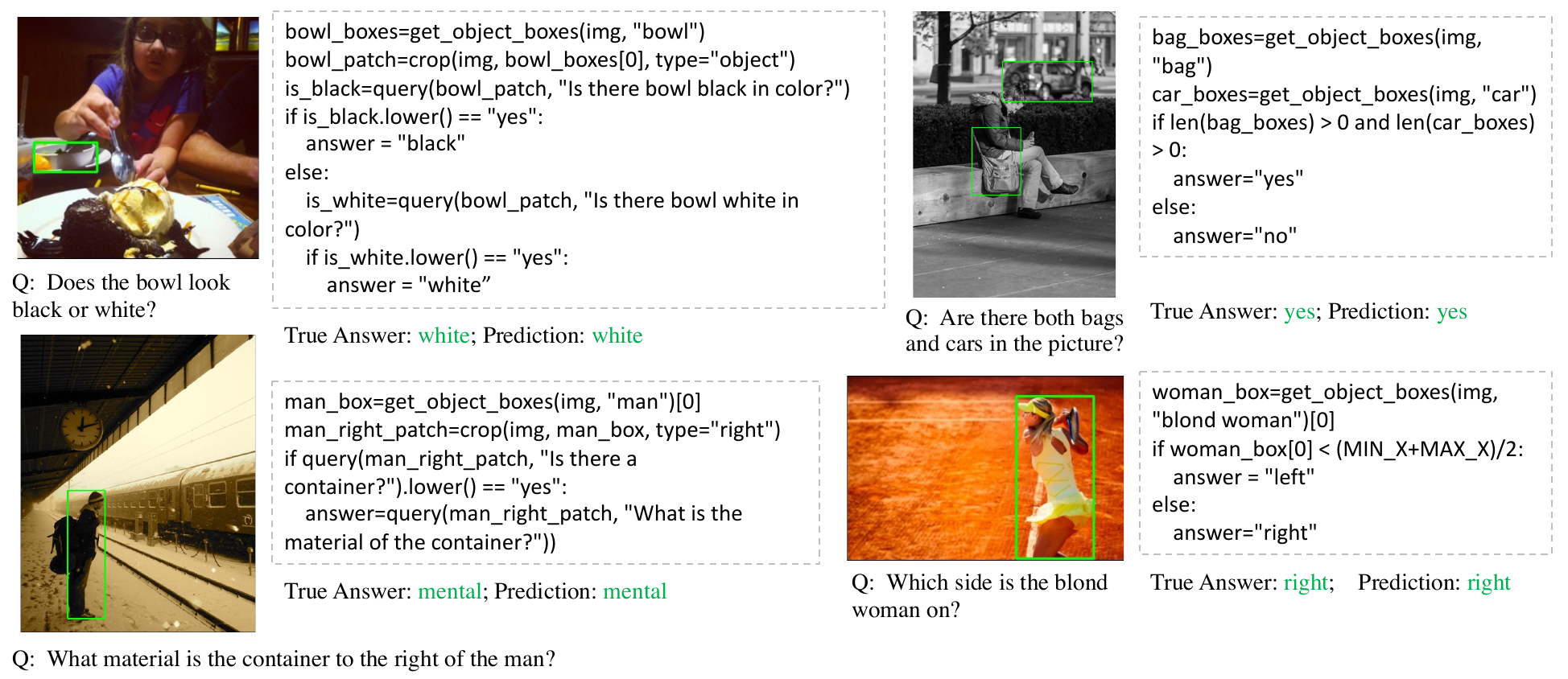}
\caption{Qualitative examples}
\label{fig:qualitative}
\end{figure*}

\noindent{\textbf{LLM.}} 
Table~\ref{tab:llm} shows the performance of PyramidCoder using different LLMs on the GQA val2000 dataset. Besides the results listed in Table~\ref{tab:sota}, we also include results using GPT-3.5, a private general LLM, for code generation to measure its performance.  
The results summarized in Table~\ref{tab:llm} demonstrate that PyramidCoder consistently improves performance across diverse LLMs.
This consistency highlights that PyramidCoder is independent of the employed LLM, as stronger LLMs usually lead to better performance. Remarkably, PyramidCoder exhibits versatility by performing well with both closed and open-source LLMs, as well as with models designed for general and code-specific tasks. 

\noindent{\textbf{Ablation study.}} 
The ablation study is conducted on the sampled GQA and VQAv2 datasets, and the results are summarized in Table~\ref{tab:ablation}. The first row shows the performance of the complete PyramidCoder. To show the significance of each component, the query rephraser, code generator, and answer aggregator are individually omitted from the whole framework. The results clearly indicate that the exclusion of any of these modules leads to a noticeable decrease in the VQA accuracy across both datasets.

\noindent{\textbf{Qualitative examples.}}
Figure~\ref{fig:qualitative} illustrates the diversified code generation capabilities of PyramidCoder. When confronted with queries of distinct content, the generated code exhibits varying levels of complexity and employs different APIs to construct disparate code solutions.

\section{Conclusion}

In this paper, we proposed PyramidCoder, a prompting framework designed for compositional visual question answering. This framework operates through query rephrasing, code candidate generation, and answer aggregation facilitated by three distinct modules. 
A key feature of this framework is its use of a single frozen LLM and predefined prompts at each level, eliminating the need for additional training and ensuring flexibility across different LLM architectures. Experimental evaluations conducted on three VQA datasets demonstrate its efficacy.

\section{Acknowledgements}
This work was supported by JSPS KAKENHI Grant Number JP23H00490 and NEDO JPNP18002.
\bibliographystyle{IEEEbib}
\bibliography{refs.bib}

\begin{thebibliography}{10}

\bibitem{Hudson2019GQADataset}
D.~A. Hudson and C.~D. Manning,
\newblock ``Gqa: A new dataset for real-world visual reasoning and compositional question answering,''
\newblock in {\em Proc. IEEE/CVF International Conference on Computer Vision (ICCV)}, 2019.

\bibitem{goyal2017vqav2}
Y.~Goyal, T.~Khot, et~al.,
\newblock ``Making the v in vqa matter: Elevating the role of image understanding in visual question answering,''
\newblock in {\em Proc. IEEE/CVF Conference on Computer Vision and Pattern Recognition (CVPR)}, 2017, pp. 6904--6913.

\bibitem{suhr2019nlvr2}
A.~Suhr, S.~Zhou, A.~Zhang, I.~Zhang, H.~Bai, and Y.~Artzi,
\newblock ``A corpus for reasoning about natural language grounded in photographs,''
\newblock in {\em Proc. Annual Meeting of the Association for Computational Linguistics (ACL)}, 2019, pp. 6418--6428.

\bibitem{Li2022GLIP}
L.~H. Li, P.~Zhang, H.~Zhang, J.~Yang, C.~Li, Y.~Zhong, L.~Wang, L.~Yuan, L.~Zhang, J.-N. Hwang, et~al.,
\newblock ``Grounded language-image pre-training,''
\newblock in {\em Proc. IEEE/CVF Conference on Computer Vision and Pattern Recognition (CVPR)}, 2022.

\bibitem{tiong2022pnpvqa}
A.~M.~H. Tiong, J.~Li, B.~Li, S.~Savarese, et~al.,
\newblock ``Plug-and-play vqa: Zero-shot vqa by conjoining large pre-trained models with zero training,''
\newblock in {\em Proc. Findings of Empirical Methods in Natural Language Processing (EMNLP Findings)}, 2022.

\bibitem{Suris2023ViperGPT}
D.~Sur\'is, S.~Menon, and C.~Vondrick,
\newblock ``{ViperGPT}: Visual inference via python execution for reasoning,''
\newblock {\em arXiv:2303.08128}, 2023.

\bibitem{Gupta2022VisProg}
T.~Gupta and A.~Kembhavi,
\newblock ``Visual programming: Compositional visual reasoning without training,''
\newblock in {\em Proc. IEEE/CVF Conference on Computer Vision and Pattern Recognition (CVPR)}, 2022.

\bibitem{Subramanian2023CodeVQA}
S.~Subramanian, M.~Narasimhan, et~al.,
\newblock ``Modular visual question answering via code generation,''
\newblock in {\em Proc. Annual Meeting of the Association for Computational Linguistics (ACL)}, 2023.

\bibitem{Wei2022ChainOfThought_CoT}
J.~Wei, X.~Wang, D.~Schuurmans, et~al.,
\newblock ``Chain-of-thought prompting elicits reasoning in large language models,''
\newblock in {\em Proc. Annual Conference on Neural Information Processing Systems (NeurIPS)}, 2022.

\bibitem{Yao2023TreeOfThoughts_ToT}
S.~Yao, D.~Yu, J.~Zhao, et~al.,
\newblock ``{Tree of Thoughts}: Deliberate problem solving with large language models,''
\newblock in {\em Proc. Annual Conference on Neural Information Processing Systems (NeurIPS)}, 2023.

\bibitem{malinowski2015ask}
M.~Malinowski, M.~Rohrbach, and M.~Fritz,
\newblock ``Ask your neurons: A neural-based approach to answering questions about images,''
\newblock in {\em Proc. IEEE/CVF International Conference on Computer Vision (ICCV)}, 2015, pp. 1--9.

\bibitem{huang2021diagnostic}
Z.~Huang, H.~Zhu, Y.~Sun, et~al.,
\newblock ``A diagnostic study of visual question answering with analogical reasoning,''
\newblock in {\em Proc. IEEE International Conference on Image Processing (ICIP)}, 2021, pp. 2463--2467.

\bibitem{zhang2022context}
H.~Zhang and W.~Wu,
\newblock ``Context relation fusion model for visual question answering,''
\newblock in {\em Proc. IEEE International Conference on Image Processing (ICIP)}, 2022, pp. 2112--2116.

\bibitem{yang2016stackedattention}
Z.~Yang, X.~He, J.~Gao, et~al.,
\newblock ``Stacked attention networks for image question answering,''
\newblock in {\em Proc. IEEE/CVF Conference on Computer Vision and Pattern Recognition (CVPR)}, 2016.

\bibitem{anderson2018bottomupandtopdown}
P.~Anderson, X.~He, C.~Buehler, et~al.,
\newblock ``Bottom-up and top-down attention for image captioning and visual question answering,''
\newblock in {\em Proc. IEEE/CVF Conference on Computer Vision and Pattern Recognition (CVPR)}, 2018, pp. 6077--6086.

\bibitem{sarkar2022grad}
A.~Sarkar and M.~Rahnemoonfar,
\newblock ``{Grad-CAM} aware supervised attention for visual question answering for post-disaster damage assessment,''
\newblock in {\em Proc. IEEE International Conference on Image Processing (ICIP)}, 2022, pp. 3783--3787.

\bibitem{wu2022ques}
X.~Wu, J.~Lu, Z.~Li, and F.~Xiong,
\newblock ``Ques-to-visual guided visual question answering,''
\newblock in {\em Proc. IEEE International Conference on Image Processing (ICIP)}, 2022, pp. 4193--4197.

\bibitem{Shen_2023_WACV}
R.~Shen, N.~Inoue, and K.~Shinoda,
\newblock ``Text-guided object detector for multi-modal video question answering,''
\newblock in {\em Proc. IEEE Conference on Winter Conference on Applications of Computer Vision (WACV)}, 2023, pp. 1032--1042.

\bibitem{li2021unsupervised}
L.~H. Li, H.~You, Z.~Wang, et~al.,
\newblock ``Unsupervised vision-and-language pre-training without parallel images and captions,''
\newblock in {\em Proc. Annual Conference of the North American Chapter of the Association for Computational Linguistics (NAACL)}, 2021.

\bibitem{le2021vttransformer}
T.~Le, H.~T. Nguyen, and M.~Le~Nguyen,
\newblock ``Vision and text transformer for predicting answerability on visual question answering,''
\newblock in {\em Proc. IEEE International Conference on Image Processing (ICIP)}, 2021, pp. 934--938.

\bibitem{brown2020gpt3}
T.~Brown, B.~Mann, N.~Ryder, et~al.,
\newblock ``Language models are few-shot learners,''
\newblock in {\em Proc. Annual Conference on Neural Information Processing Systems (NeurIPS)}, 2020, pp. 1877--1901.

\bibitem{touvron2023llama}
H.~Touvron, L.~Martin, K.~Stone, et~al.,
\newblock ``Llama 2: Open foundation and fine-tuned chat models,''
\newblock {\em arXiv preprint arXiv:2307.09288}, 2023.

\bibitem{Chen2021Codex}
M.~Chen, J.~Tworek, et~al.,
\newblock ``Evaluating large language models trained on code,''
\newblock {\em arXiv2107.03374}, 2021.

\bibitem{Li2023StarCoder}
R.~Li et~al.,
\newblock ``Starcoder: may the source be with you!,''
\newblock {\em arXiv:2305.06161}, 2023.

\bibitem{roziere2023codellama}
B.~Roziere, J.~Gehring, F.~Gloeckle, et~al.,
\newblock ``Code llama: Open foundation models for code,''
\newblock {\em arXiv preprint arXiv:2308.12950}, 2023.

\bibitem{team2023gemini}
G.~Team, R.~Anil, et~al.,
\newblock ``Gemini: a family of highly capable multimodal models,''
\newblock {\em arXiv preprint arXiv:2312.11805}, 2023.

\bibitem{Wang2023SelfConsistency}
X.~Wang, J.~Wei, D.~Schuurmans, et~al.,
\newblock ``Self-consistency improves chain of thought reasoning in language models,''
\newblock in {\em Proc. International Conference on Learning Representations (ICLR)}, 2023.

\bibitem{Zhang2023AutomaticCoT}
Z.~Zhang, A.~Zhang, M.~Li, and A.~Smola,
\newblock ``Automatic chain of thought prompting in large language models,''
\newblock in {\em Proc. International Conference on Learning Representations (ICLR)}, 2023.

\bibitem{Yao2022ReAct}
S.~Yao, J.~Zhao, D.~Yu, et~al.,
\newblock ``React: Synergizing reasoning and acting in language models,''
\newblock 2023.

\bibitem{Johnson2017Inferring}
J.~Johnson, B.~Hariharan, L.~van~der Maaten, et~al.,
\newblock ``Inferring and executing programs for visual reasoning,''
\newblock in {\em Proc. IEEE/CVF International Conference on Computer Vision (ICCV)}, 2017.

\bibitem{liu2023groundingdino}
S.~Liu, Z.~Zeng, T.~Ren, et~al.,
\newblock ``{Grounding DINO}: Marrying {DINO} with grounded pre-training for open-set object detection,''
\newblock {\em arXiv preprint arXiv:2303.05499}, 2023.

\bibitem{li2022blip}
J.~Li, D.~Li, C.~Xiong, and S.~Hoi,
\newblock ``{BLIP}: Bootstrapping language-image pre-training for unified vision-language understanding and generation,''
\newblock in {\em Proc. International Conference on Machine Learning (ICML)}, 2022, pp. 12888--12900.

\end{thebibliography}
\end{document}